\documentclass[dvipsnames]{article}

\usepackage[final]{text}

\usepackage[utf8]{inputenc} %
\usepackage[T1]{fontenc}    %
\usepackage{url}            %
\usepackage{booktabs}       %
\usepackage{amsfonts}       %
\usepackage{nicefrac}       %
\usepackage{booktabs} 
\usepackage{multirow}
\usepackage{wrapfig}
\usepackage{amssymb}
\usepackage{epigraph}
\usepackage{makecell}
\usepackage{listings}
\usepackage{adjustbox}
\usepackage[labelfont=bf]{caption} %
\usepackage{forest}
\usepackage{tikz}
\usetikzlibrary{tikzmark,calc}
\usepackage[ruled,vlined]{algorithm2e}
\usepackage{color, soul}
\usepackage{microtype}      %
\usepackage{xcolor}         %
\usepackage{geometry}
\usepackage{times}
\usepackage{ragged2e}
\usepackage{amsmath}
\usepackage{bm}
\usepackage{latexsym}
\usepackage{subfigure}
\usepackage{graphicx}
\usepackage{float}
\usepackage{longtable}
\usepackage{booktabs} 
\usepackage{multirow}
\usepackage{amssymb}
\usepackage{longtable}
\usepackage{tabu}
\usepackage{bbding}
\usepackage{awesomebox}
\usepackage{bbding}
\usepackage[most,breakable]{tcolorbox}
\usepackage{etoolbox}
\usepackage{fancyhdr}
\usepackage{lipsum}
\usepackage{CJKutf8}
\usepackage{pdfpages}
\usepackage{multicol}
\usepackage{pgffor} 

\usepackage{pifont}

\usepackage{caption} 
\usepackage{chngcntr}

\usepackage[colorlinks, linkcolor=RoyalBlue, anchorcolor=BrickRed, citecolor=RoyalBlue, urlcolor=RoyalBlue]{hyperref}

\usepackage{cleveref}
\crefname{section}{§}{§§}
\Crefname{section}{§}{§§}

\newcommand\refsec[1]{Section~\hyperref[sec:#1]{\ref{sec:#1}}}
\newcommand\refsecs[2]{\hyperref[sec:#1]{§\ref{sec:#1}:~\textsc{#1}}, \hyperref[sec:#2]{§\ref{sec:#2}:~\textsc{#2}}}

\usepackage[tikz]{bclogo}
\definecolor{msftBlue}{RGB}{0,164,239}
\definecolor{msftGreen}{RGB}{127,186,0}
\definecolor{msftYello}{RGB}{255,185,0}
\definecolor{msftBlack}{RGB}{0,0,0}

\newtcolorbox{myboxnote}[1][]{
  breakable,
  title=#1,
  colback=cyan!0,
  colbacktitle=cyan!0,
  coltitle=black,
  fonttitle=\bfseries,
  bottomrule=0pt,
  toprule=0pt,
  leftrule=1.5pt,
  rightrule=1.5pt,
  titlerule=0pt,
  arc=0pt,
  outer arc=0pt,
  colframe=lightgray,
}
\usepackage{mdframed}

\definecolor{academicblue}{RGB}{54, 95, 145}
\usepackage{multirow}

\newtcbox{\smybox}[1][red]{on line,
arc=1pt,colback=#1!10!white,colframe=#1!100!black,
before upper={\rule[-3pt]{0pt}{10pt}},
boxsep=0pt,left=6pt,right=6pt,top=2pt,bottom=0pt,boxrule=0pt,leftrule=1pt,rightrule=1pt}

\newenvironment{itemize*}%
 {\leftmargini=20pt\begin{itemize}%
  \setlength{\itemsep}{3pt}%
  \setlength{\parskip}{0pt}%
  }%
 {\end{itemize}}
\newenvironment{enumerate*}%
 {\begin{enumerate}%
  \setlength{\itemsep}{0pt}%
  \setlength{\parskip}{0pt}}%
 {\end{enumerate}}

\usepackage{fancyhdr} %
\usepackage{blindtext} %
\usepackage{ulem}
\usepackage{fontawesome5} %

\usepackage{xparse}

\usepackage{colortbl}

\newcommand{\hypbox}[2]{%
\begin{tcolorbox}[colback=white!98!black,colframe=white!30!black,boxsep=1.1pt,top=6.75pt]%
\vspace{1.75pt}%
\textbf{#1}\\[-0.575em]
\noindent\makebox[\textwidth]{\rule{\textwidth}{0.4pt}}
\\[0.25em]
#2
\end{tcolorbox}
}

\title{
  Towards Building Specialized Generalist AI \\
  with System 1 and System 2 Fusion
}

\author{Kaiyan Zhang$^{1}$\thanks{Equal contributions} \quad \ Biqing Qi$^{1*}$ \quad Bowen Zhou$^{1,2}$\thanks{Corresponding author}\vspace{0.03in}\\
$^1$Tsinghua University \quad $^2$Shanghai AI Laboratory\\
\texttt{zhoubowen@tsinghua.edu.cn}\\
}

\begin{document}

\maketitle

\vspace{-0.5cm}

\begin{abstract}
In this perspective paper, we introduce the concept of Specialized Generalist Artificial Intelligence (SGAI or simply SGI) as a crucial milestone toward Artificial General Intelligence (AGI).
Compared to directly scaling general abilities, SGI is defined as AI that specializes in at least one task, surpassing human experts, while also retaining general abilities. This fusion path enables SGI to rapidly achieve high-value areas.
We categorize SGI into three stages based on the level of mastery over professional skills and generality performance. Additionally, we discuss the necessity of SGI in addressing issues associated with large language models, such as their insufficient generality, specialized capabilities, uncertainty in innovation, and practical applications.
Furthermore, we propose a conceptual framework for developing SGI that integrates the strengths of Systems 1 and 2 cognitive processing~\footnote{Described by psychologist Daniel Kahneman's book ``Thinking, Fast and Slow'', System 1 is fast, automatic, and intuitive thinking, while System 2 is slow, deliberate, and analytical thinking.}. This framework comprises three layers and four key components, which focus on enhancing individual abilities and facilitating collaborative evolution.
We conclude by summarizing the potential challenges and suggesting future directions.
We hope that the proposed SGI will provide insights into further research and applications towards achieving AGI.
\end{abstract}

\begin{figure*}[htbp]
\vskip 0.2in
\begin{center}
\centerline{
\includegraphics[width=0.86\textwidth]{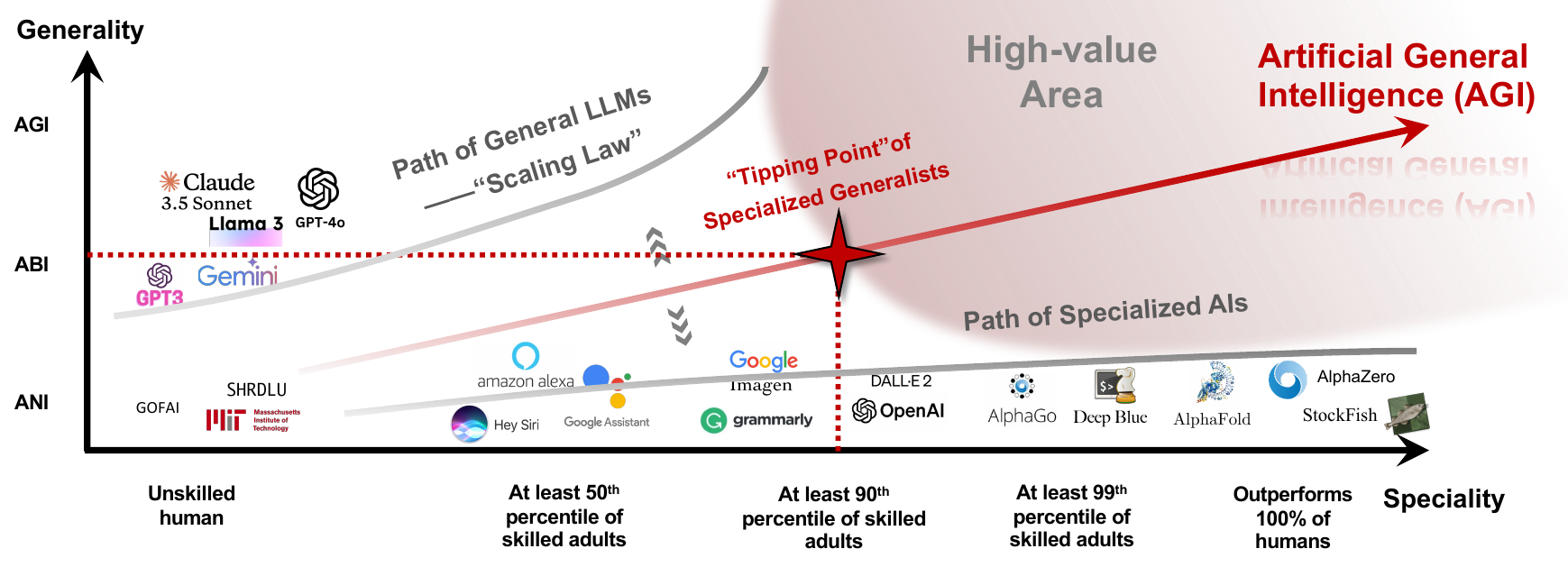}
}
\caption{
The role of Specialized Generalist Intelligence (SGI) is a crucial milestone toward Artificial General Intelligence (AGI). The implementation path of AGI encompasses two dimensions: speciality and generality. The development of speciality, as detailed in “Levels of AGI”~\citep{morris2024position}, is often compared with human intelligence.
The differences in generality lie in the number of skills held, which also depend on distinct learning paradigms.
While the ``Scaling Law'' has led to significant improvements in the generality of LLMs, progress in speciality remains extremely slow.
Upon reviewing the development of technology, a high-value area emerges for AGI applications that require models to possess both robust generality and adequate specialty. This optimal balance point is referred to as the ``tipping point'' for specialized generalists.
}
\label{fig:SGI}
\end{center}
\vskip -0.2in
\end{figure*}

\newpage
{
  \hypersetup{linkcolor=RoyalBlue, linktoc=page}
  \tableofcontents
}

\newpage

\section{Introduction}

Artificial intelligence (AI)~\citep{mccarthy2006proposal} has a long history, dating back to the Dartmouth Summer Research Project~\footnote{\url{https://en.wikipedia.org/wiki/Dartmouth_workshop}} in 1956, which aimed to complete intellectual tasks automatically without human intervention.
Over the past 60 years, various notable AI applications, such as Deep Blue~\citep{campbell2002deep}, Watson~\citep{wiki:IBM_Watson}, AlphaGo~\citep{silver2017mastering}, and AlphaFold~\citep{jumper2021highly}, have surpassed almost all skilled adults in specialized tasks.
Recently, the emergence of Large Language Models (LLMs) such as GPT-4~\citep{achiam2023gpt}, Gemini 1.5~\citep{reid2024gemini}, Claude~\citep{claude} and Llama 3~\citep{llama3modelcard} has significantly accelerated AI research and applications in general tasks.

According to the ``Scaling Law''~\citep{kaplan2020scaling,hoffmann2022training}, LLMs with massive parameters trained on extremely large corpora exhibit exceptionally high intelligence compared to earlier AIs~\citep{krizhevsky2012imagenet,sutskever2014sequence,he2016deep,devlin2019bert}.
LLMs and LLM-based agents demonstrate advanced capabilities in instruction following~\citep{alpaca,ding2023enhancing,xu2023wizardlm}, programming~\citep{qian2023communicative,yang2024sweagent,shao2024deepseekmath}, mathematics~\citep{lightman2023let,trinh2024solving}, and various downstream applications.
These models have achieved human-level understanding in various academic benchmarks and have even surpassed human experts in specific domains, such as medical~\citep{nori2023can,saab2024capabilities} and financial~\citep{kim2024financial} tasks.

However, LLMs still fail to outperform humans on most tasks, especially those requiring reasoning and deliberation~\citep{huang2023large,wan2024b}, which is a characteristic of Artificial General Intelligence (AGI).
Broadly defined, AGI refers to machines capable of performing any intellectual task that a human can do~\citep{feng2024far}.
According to recent definitions of AGI levels~\citep{morris2024position}, powerful LLM-based assistants like ChatGPT and Gemini can be regarded as emerging AGI, comparable to or somewhat better than an unskilled human across a wide range of non-physical tasks.
Although ``sparks'' of AGI are already present in the latest LLMs~\citep{bubeck2023sparks} in terms of general ability, they are still far from achieving real AGI.
The next step in the development of AGI from generality is competent AGI~\citep{morris2024position}, which should achieve performance at least at the 50th percentile of skilled adults compared to emerging AGI.

In practice, the definition of AGI is challenging to discern in real-world deployments and high-value applications. The applications of the latest LLMs in most real-world scenarios are still experimental and do not yet produce high-value innovations.
This raises a critical question:

\textit{What should be the next step when optimizing LLMs towards AGI from a practical standpoint?}

In contrast to the gradual advancement of scaling general capabilities towards AGI, we aim to prioritize a more expedited approach. This path focuses on achieving a breakthrough in at least one specialized capability while maintaining essential general capabilities, thereby defining the core of Specialized Generalist Intelligence (SGI). Specifically, our goal is to create a self-reinforcing AGI acceleration path by generating high-efficiency data from significant feedback effects.
\textbf{In this position paper, we argue that it is essential to strive for the achievement of SGI at this juncture.} From the perspective of intelligence relative to skilled adults, previous AIs specialized in one task, like AlphaGo and AlphaFold, should be regarded as specialists, achieving performance better than 90\% of skilled adults (human experts). Meanwhile, AGI is a generalist, possessing higher specialized abilities across various tasks compared to human experts in almost any specific domain.
\textbf{On the other hand, SGI refers to AI systems with intelligence specialized in at least one task, surpassing human experts, while also maintaining general abilities superior to those of an unskilled human across nearly any task.}
In summary, in the context of SGI system development, we envision that such systems should possess the critical abilities to continuously learn and adapt to new tasks, autonomously uncover novel knowledge, and optimize their objectives in alignment with human values.

In the following sections, we first introduce the concept of specialized generalist AI in $\S$~\ref{sec:what_sgi}, where we define it within the context of AGI levels~\citep{morris2024position}.
Then, we discuss the necessity of specialized generalists in $\S$~\ref{sec:why_sgi}.
Next, we present a conceptual framework from the lens of System 1 and System 2 in $\S$~\ref{sec:framework}, which encompasses four components aimed at enhancing the capabilities of System 1 and System 2 and their interactions.
Finally, we discuss challenges and future directions for achieving specialized generalists in $\S$~\ref{sec:challenge}.

\section{What is Specialized Generalist AI?}
\label{sec:what_sgi}

\hypbox{Takeaway Message}{
As significant milestones toward expert-level AGI, Specialized Generalist AI (SGI) possesses specialized abilities in specific tasks, surpassing 90\% of human experts, while also maintaining a general capability across a wide range of tasks at least comparable to an unskilled human’s abilities.
}

In this section, we describe what specialized generalist AI is.
We first introduce the three stages of AI in $\S$~\ref{sec:what_sgi_ai_stages} and then identify the current state of AI especially LLMs in $\S$~\ref{sec:what_sgi_ai_state}.
Finally, we define the position of SGI and outline its five stages in $\S$~\ref{sec:what_sgi_def}.

\subsection{Three Stages of AI}
\label{sec:what_sgi_ai_stages}

Reviewing the development process of AI~\citep{muggleton2014alan,haenlein2019brief}, we can divide it into three main stages based on the scope of intelligence~\citep{IBM_Watson_ABI}:

\begin{itemize}
    \item \textbf{Stage 1: Artificial Narrow Intelligence (ANI).}
    ANI primarily relies on supervised learning~\citep{krizhevsky2012imagenet,russakovsky2015imagenet} with large amounts of manually labeled data.
    The task scope of ANI is narrow, and new AI models need to be trained for new tasks.
    During this period, we know exactly what AI can and cannot do, indicating that AIs lack the ability to generalize to new tasks.
    Generally speaking, efforts on ANI were largely completed by 2016, just before the emergence of transformers and self-supervised pre-training like ELMO~\citep{peters2018deep}, BERT~\citep{devlin2019bert}, and GPT~\citep{radford2018improving}.
    \item \textbf{Stage 2: Artificial Broad Intelligence (ABI).}
    The most significant feature of ABI is the self-supervision learning paradigm, which reduces the need for extensive explicit teaching for specific tasks.
    The learning method follows an end-to-end approach based on scalable architectures like attention-based~\citep{lin2017structured} transformers~\citep{vaswani2017attention}, enabling AI to decompose and complete multiple tasks independently.
    Using the next token prediction learning objective~\citep{Radford2019LanguageMA,brown2020language}, there is a paradigm shift from discriminative and classification-based AIs to generative assistants~\citep{zhou2024generative}.
    This stage encompasses pre-trained models and large language models and is currently ongoing.
    \item \textbf{Stage 3: Artificial General Intelligence (AGI).} 
    The AGI period has not yet arrived. AGI will be smarter than humans and continuously learn to become more proficient in any task.
    The learning paradigm of AGI should be independent and self-developed, meaning fully autonomous learning without human intervention.
    At this stage, there will be unexpected risks and uncontrollable factors as AGI’s abilities increase.
    Therefore, governance and regulation are essential, motivating super-alignment research~\citep{burns2023weak}.
\end{itemize}

\begin{table*}[htbp]
\caption{
Based on the ``Levels of AGI'' proposed by~\citet{morris2024position}, we introduce a new dimension, ``Broad,'' for SGI, positioned between ``Narrow'' and ``General.'' SGI encompasses three stages, reflecting the number of skills at the expert level and performance on general tasks. This represents a leveled, matrixed approach to classifying systems on the path to AGI, based on the depth (performance) and breadth (generality) of capabilities. The placement of example systems within this matrix is approximate.
The \colorbox{red!20}{red cells} in the table represent stages and levels related to SGI, which also outline the proposed path in this paper toward Expert AGI from Emerging AGI.
The \colorbox{blue!20}{blue cells} represent the beginning of SGI, respectively.
We categorize the status of Levels 1 and 2 within broad AI as \textcolor{gray}{\textbf{Intermediate State}}, which, while valuable (like Siri and Alexa in narrow tasks), struggles to produce high value.
}
\label{tab:levels_of_AGI}
\begin{center}
\begin{small}
\begin{tabular}{p{0.22\textwidth}|p{0.22\textwidth}|p{0.21\textwidth}|p{0.21\textwidth}}
    \toprule
    \textbf{Performance} (rows) x \newline \textbf{Generality} (columns) & \textbf{Narrow} \newline \textit{clearly scoped task or set of tasks} & 
    \textbf{Broad} \newline \textit{
    diverse and complex set of tasks across multiple domains
    } &
    \textbf{General} \newline \textit{wide range of non-physical tasks, including metacognitive tasks like learning new skills} \\ 
    \midrule
    \textbf{Level 0: No AI} & \textbf{Narrow Non-AI} \newline calculator software; compiler & \textbf{Broad Non-AI}  & \textbf{General Non-AI} \newline human-in-the-loop computing, e.g., Amazon Mechanical Turk  
    \\ \midrule
    \textbf{Level 1: Emerging} \newline \textit{equal to or somewhat better than an unskilled human} & \textbf{Emerging Narrow AI} \newline GOFAI \citep{Boden_2014}; simple rule-based systems, e.g., SHRDLU \citep{shrdlu} & 
    \textcolor{gray}{\textbf{Intermediate state}}
    & \cellcolor{blue!20} \textbf{Emerging AGI} \newline ChatGPT \citep{openai2023gpt4}, Bard \citep{anil2023palm}, Llama 2 \citep{touvron2023llama}, Gemini \citep{geminiBlog} \\ \midrule
    \textbf{Level 2: Competent} \newline \textit{at least 50th percentile of skilled adults} & \textbf{Competent Narrow AI} \newline  toxicity detectors such as Jigsaw \citep{das2022toxic}; Smart Speakers such as Siri \citep{siri}, Alexa \citep{alexa}, or Google Assistant \citep{gasst}; VQA systems such as PaLI \citep{chen2023pali}; Watson \citep{watson}; SOTA LLMs for a subset of tasks (e.g., short essay writing, simple coding) & 
    \textcolor{gray}{\textbf{Intermediate state}}
    & \cellcolor{gray!20} \textbf{Competent AGI} \newline not yet achieved \\ \midrule
    \textbf{Level 3: Expert} \newline \textit{at least 90th percentile of skilled adults} & \textbf{Expert Narrow AI} \newline spelling \& grammar checkers such as Grammarly \citep{grammarly}; generative image models such as Imagen \citep{saharia2022photorealistic} or Dall-E 2 \citep{dalle2} & \cellcolor{red!20} 
    \textbf{Specialized Generalists \newline Intelligence (SGI)} \newline
    \textit{\underline{Stage 1: Emerging SGI}}\newline e.g., MedPrompt~\citep{nori2023can},MedGemini~\citep{saab2024capabilities} \newline
    \textit{\underline{Stage 2: Competent SGI}}\newline not yet achieved \newline
    \textit{\underline{Stage 3: Expert SGI}} \newline not yet achieved \newline
    & \cellcolor{gray!20} \textbf{Expert AGI} \newline not yet achieved  \\ \midrule
    \textbf{Level 4: Virtuoso} \newline \textit{at least 99th percentile of skilled adults} & \textbf{Virtuoso Narrow AI} \newline Deep Blue \citep{deepblue}, AlphaGo \citep{alphago, alphagoRL} & \multicolumn{2}{c}{\cellcolor{gray!20} {\textbf{Virtuoso AGI} \quad \newline not yet achieved}}  \\ \midrule
    \textbf{Level 5: Superhuman} \newline \textit{outperforms 100\% of humans} & \textbf{Superhuman Narrow AI} \newline AlphaFold \citep{alphafold1,alphafold2}, AlphaZero \citep{alphazero}, StockFish \citep{stockfish} 
    & \multicolumn{2}{c}{\cellcolor{gray!20} {\textbf{Artificial Superintelligence (ASI)} \quad \newline not yet achieved }} \\ \bottomrule
    \end{tabular}
\end{small}
\end{center}
\end{table*}

\subsection{The Current Stage of AI}
\label{sec:what_sgi_ai_state}

Overall, current AI, especially LLMs, still resides between ABI and AGI, making it challenging to pinpoint the exact stage.
To quantify AI’s abilities and performance on tasks, \citet{morris2024position} proposed levels of AGI to outline different stages of AGI in comparison to human performance. As shown in Table~\ref{tab:levels_of_AGI}, there are six levels, ranging from no AI (Level 0) to superhuman AI (Level 5), which outperforms 100\% of humans.
The remaining four levels, each skilled in any task, are shown as follows:
\begin{itemize}
    \item \textbf{Level 1:} Represents an emerging level, somewhat better than an unskilled human.
    \item \textbf{Level 2:} Competent level, reaching at least the 50th percentile of skilled adults.
    \item \textbf{Level 3:} Expert level, reaching at least the 90th percentile of skilled adults.
    \item \textbf{Level 4:} Virtuoso level, reaching at least the 99th percentile of skilled adults.
\end{itemize}

Based on this categorization, current LLM-based AIs like ChatGPT~\citep{openai2023gpt4}, Llama~\citep{llama3modelcard} and Gemini~\citep{reid2024gemini} belong to emerging AGI (i.e., Level 1), which is equal to or somewhat better than an unskilled human across a wide range of non-physical tasks.
In addition to performance comparisons with humans, AGI should also encompass preliminary metacognitive tasks~\citep{wang2023metacognitive,didolkar2024metacognitive}, such as learning new skills~\citep{lake2023human}.

\subsection{Definition, Capabilities and Stages of SGI}
\label{sec:what_sgi_def}

This section defines Specialized Generalist AI (SGI) and delineates its developmental stages, categorized by the number of specialized abilities depicted in Figure~\ref{fig:SGI}.

\textbf{Definition of SGI.}
SGI builds upon advanced LLMs (i.e., emerging AGI, expectedly endowing it with capabilities that surpass those of over 90\% of human experts in specialized domains.
Unlike domain-specific ANI, SGI exhibits enhanced general capabilities while retaining deep expertise in specific tasks relative to emerging AGI.
The development of SGI spans two dimensions: the acquisition of a broad range of expert-level specialized abilities, and the enhancement of general capabilities across diverse textual tasks. The first dimension involves advancing from Level 3 to Level 5 AGI, leveraging emerging AGI technologies across as many tasks as possible. The second dimension focuses on the continuous improvement of general capabilities in AGI on any task.
Through targeted skill optimization, it's expected to surpass the general Scaling Law.

\textbf{Capabilities of SGI.}
Specifically, in the development of SGI systems, we anticipate that these systems will exhibit the following three core capabilities:
\begin{itemize}
    \item \textbf{Task Streaming Learning Ability:} SGI systems should possess cross-domain, long-term, and continuous learning capacities~\citep{lake2023human,wang2023incorporating,qi2024interactive}. This would enable them to handle an infinite number of tasks, flexibly iterating and generalizing from new task learnings in a task flow context. This capability not only underscores the broad applicability of SGI systems but also their potential for rapid autonomous learning.
    \item \textbf{Autonomous Discovery Capability:} SGI systems should be able to autonomously uncover new knowledge and identify novel tasks. They should also develop a comprehensive self-assessment and metrics system to ensure the continuous emergence of new knowledge and discoveries, driving sustained progress in science and technology~\citep{volk2023alphaflow,boiko2023autonomous}. This capability significantly enhances the innovative and autonomous exploration capacities of SGI systems.
    \item \textbf{Value-Aligned Optimization Ability:} SGI systems should be capable of establishing a task goal system aligned with human values~\citep{yuan2022situ}. This alignment would enable them to integrate more closely with the real physical environment and better serve the genuine needs of human society. Cultivating this capability not only improves the practical applicability of SGI systems but also augments their positive contributions to human social development.
\end{itemize}

\textbf{Stages of SGI.}
Drawing from the ``Levels of AGI'' described by~\citet{morris2024position}, we delineate the stages of SGI based on the breadth of expert-level specialized tasks and depth of general task capabilities.
\begin{itemize}
    \item \colorbox{red!20}{\textbf{\underline{Stage 1: Emerging SGI.}}} At this initial stage, SGI achieves at least 90th percentile of skilled adults in one specific domain while maintaining general abilities across a wide range of non-physical tasks, comparable to or slightly better than those of an unskilled human. Recent works~\citep{saab2024capabilities,yang2024advancing} in biomedicine based on LLMs are approaching this stage, although their expert-level abilities still require more comprehensive evaluation.
    \item \colorbox{red!20}{\textbf{\underline{Stage 2: Competent SGI.}}} Here, SGI demonstrates competency in approximately 20\% of typical tasks and domains encountered in everyday life and work. Inspired by the Pareto Principle (80/20 rule)~\citep{dunford2014pareto}, mastering 20\% of skills across all domains enables handling nearly 80\% of typical scenarios. At this stage, SGI’s capabilities are at least on par with the median skill level of skilled adults across a broad spectrum of tasks.
    \item \colorbox{red!20}{\textbf{\underline{Stage 3: Expert SGI.}}} At this advanced stage, SGI closely approaches the capabilities of Expert AGI, proficiently handling about 90\% of tasks worldwide and ranking in the 90th percentile among skilled adults in these areas. Concurrently, SGI’s general abilities fall between the 50th and 90th percentiles of skilled adults in a variety of non-physical tasks.
    \item \colorbox{gray!20}{\textbf{Stage 4: Expert AGI.}} In this final stage, SGI transitions to Expert AGI, reaching proficiency that matches or exceeds the 90th percentile of skilled adults across a wide range of non-physical tasks.
\end{itemize}

\textbf{Relationship between Capabilities and Stages of SGI.}
As intelligent systems progress from the emerging SGI stage to the expert AGI stage, their three key capabilities exhibit a spiral-like enhancement~\footnote{\url{https://en.wikipedia.org/wiki/Spiral}}: continuous task learning ability, autonomous knowledge discovery ability, and value alignment optimization capability.
In the emerging SGI stage, systems focus on a specific domain and, through continuous task learning, achieve expert-level proficiency while maintaining a basic level of general capability. Upon entering the proficient SGI stage, the system’s autonomous discovery ability is enhanced, enabling it to master 20\% of the key skills necessary to handle 80\% of common tasks, demonstrating a broader capacity for generalization.
Progressing to the expert SGI stage, the system’s task learning and adaptation capabilities are significantly improved, allowing it to competently handle 90\% of tasks. Simultaneously, its autonomous knowledge discovery ability makes substantial progress, enabling it to independently explore and uncover new insights without relying excessively on existing human knowledge supervision.
Ultimately, upon reaching the expert AGI stage, the system’s key capabilities reach a top-tier level. Not only can it handle over 90\% of non-skill-based tasks, but it also exhibits exceptional performance in autonomous discovery and value alignment, aligning closely with human society.
In summary, the continuous enhancement of these three key capabilities is a significant hallmark of the intelligent system’s evolution from emerging to expert stages, as they mutually reinforce and collectively drive the system towards higher levels of intelligence.

Current specialized LLMs~\cite{nori2023can,saab2024capabilities} are close to Stage 1, although further efforts are still required to fully reach this stage.
The subsequent stages have not yet been achieved and require additional exploration, which is also influenced by the enhancement of general abilities.
Overall, SGI represents a critical pathway toward AGI, offering more pragmatic objectives of feasibility from the standpoint of practical applications.
SGI also serves as a bridge between emerging LLM-based intelligence and AGI.
Therefore, LLMs form a crucial backbone for achieving SGI and will be the focus of future research, as detailed in $\S$~\ref{sec:framework}.

\section{Why Specialized Generalist?}
\label{sec:why_sgi}

\hypbox{Takeaway Message}{
While advanced LLMs excel in specific benchmarks, they often struggle with nuanced reasoning and adaptability in varied contexts.
Specialized models may achieve high alignment with human instructions but frequently at the expense of creativity and innovation.
Given the same resources, specialized generalists offers a more efficient pathway toward achieving AGI, characterized by lower costs and increased speed compared to traditional ``Scaling Laws''.
}

\subsection{Generalists Are Not Sufficiently General}
\label{sec:why_sgi_generalist}
Advanced Large Language Models (LLMs) such as GPT-4, Claude-3, and Llama-3 have already surpassed unskilled human performance across various benchmarks.
For instance, the latest version of GPT-4 Omni~\footnote{\url{https://openai.com/index/hello-gpt-4o/}}, achieved an average score of 88.7 on the MMLU~\citep{hendrycks2020measuring}, a mixed collection of high school and undergraduate-level questions across various subjects, while the average score of domain experts is 89.7.
In the case of GPQA~\citep{rein2023gpqa}, a challenging dataset composed of graduate-level questions, Claude 3.5 Sonnet~\footnote{\url{https://www.anthropic.com/news/claude-3-5-sonnet}} scored an average of 67.2, which is higher than the average score of 65 achieved by domain expert PhDs. 
LLMs perform exceptionally well, often surpassing human experts based on benchmarks.

However, the lack of real reasoning ability in LLMs cannot be overlooked despite their general capabilities~\citep{berglund2023reversal,huang2023large,valmeekam2023can,stechly2023gpt}.
Despite their capabilities, LLMs still often fail to perform as well as an average unskilled human in many cases, even making unexpected commonsense mistakes~\citep{berglund2023reversal,wang2024can}.
There is now a debate that the auto-regression learning paradigm offers only a superficial understanding of the world.
Recent research also indicates that the neural network responsible for generating and analyzing language in the human brain does not govern formal reasoning, suggesting that reasoning does not necessarily require language as a medium~\citep{fedorenko2024language}.
Moreover, due to the limited knowledge embedded in language relative to the vast common sense knowledge available in the multimodal real-world~\citep{lecun2022path,feng2024far}, language models fall short as text-based world simulators~\citep{wang2024can}.
Typical LLMs-based emerging AGI possess preliminary generality but are still far from achieving Expert AGI.
They are not yet stable enough in their current general state when faced with little disturbed tasks~\citep{chen2024premise,zhang2024careful,hong2024stuck}.
These issues, likely constrained by current data, algorithms, and even architecture, remains an area of ongoing exploration.

Therefore, LLMs are not sufficiently general, as there is quite a gap between LLMs and the human experts.
Rather than solely striving for Expert AGI in any task, it is now essential to explore models that specialize in at least one expert-level domain while retaining their generality across multiple domains.

\subsection{Specialists Lack Uncertainty in Innovations}
\label{sec:why_sgi_specialist}

Alignment methods such as supervised fine-tuning and preference learning, including techniques like Reinforcement Learning from Human Feedback (RLHF)~\citep{schulman2017proximal,ouyang2022training} and Direct Preference Optimization (DPO)~\citep{rafailov2024direct}, enhance the adherence of base LLMs to human values and improve their responsiveness to instructions.
However, high levels of alignment are not always beneficial.
Recent studies indicate that diversity and creativity in LLMs diminish following alignment~\citep{franceschelli2023creativity,mohammadi2024creativity}.
For example, self-consistency~\citep{wang2022self} is an effective method to ensemble LLMs’ results, where uncertainty in answers is crucial. However, the effectiveness of self-consistency decreases in highly aligned models~\citep{tian2023just,fierro2024does,xiong2023can}.
This suggests that LLMs tend to conform to human intentions present in the instruction datasets, while their ability to generalize to out-of-domain contexts appears to be impaired.

In fact, uncertainty~\citep{gawlikowski2023survey,kong2023uncertainty} is crucial for innovations such as scientific discovery~\citep{park2023papers}, where previous studies have focused on enhancing uncertainty in LLMs' explorations to generate novel hypotheses~\citep{zhang2021leveraging,qi2023large}.
High uncertainty in LLMs can foster the production of more diverse candidates~\citep{boiko2023autonomous,romera2024mathematical}, among which the ground truth is more likely to be found.
From this perspective, fusing generalization in specialists is indispensable, a subject that is gradually being studied in fields such as coding~\citep{zhu2024deepseek}, medicine~\citep{Zhang2024UltraMedicalBS}, and chemistry~\citep{zhang2024chemllm}.
Yet, there remains a significant gap in achieving a trade-off between specialization and generalization, and in validating the effectiveness of specialized generalists. Furthermore, we must not overlook the importance of generalists while developing specialized LLMs in target domains.

\section{Conceptual Framework}
\label{sec:framework}
In previous sections, we have outlined what Specialized Generalist Intelligence (SGI) is and why it is necessary. As mentioned in $\S$~\ref{sec:what_sgi_def}, we believe SGI should possess three core capabilities. To achieve these capabilities, we propose a conceptual framework based on the dual-process theory of Systems 1 and 2. This framework includes three layers and four key components designed to integrate the abilities of specialists and generalists in applications, summarizing their contributions to the development of specialized generalists. We provide preliminary background on Systems 1 and 2 in $\S$~\ref{sec:framework_sys1_and_2}. Subsequently, we introduce:
Layer 1 in $\S$~\ref{sec:framework_layer1} to enhance foundational abilities in generality and specialty, including the enhancement of System 2 capabilities in $\S$~\ref{sec:framework_layer1_component1} and System 1 capabilities in $\S$~\ref{sec:framework_layer1_component2};
Layer 2 in $\S$~\ref{sec:framework_layer2} to implement collaborative fusion between Systems 1 and 2;
Layer 3 in $\S$~\ref{sec:framework_layer3} to achieve interactive self-evolving, including systematic continual task learning.

\begin{figure*}[htbp]
\vskip 0.2in
\begin{center}
    \centerline{\includegraphics[width=0.92\textwidth]{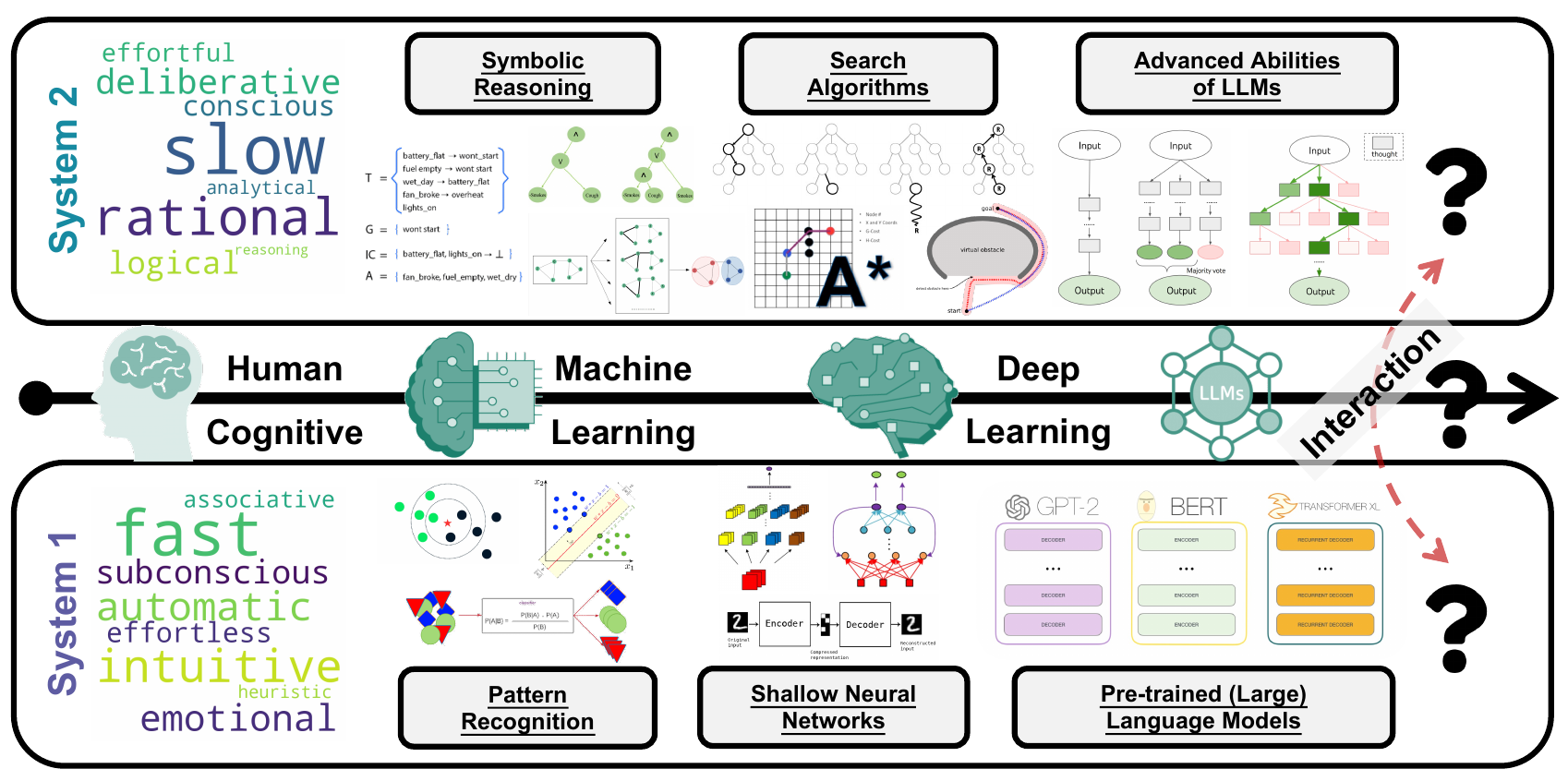}}
  \caption{
  This image presents the features of Systems 1 and 2 in human cognition, and illustrates their progression from machine learning and deep learning to LLMs in artificial intelligence, emphasizing their evolutionary journey and applications over time.
  }
  \label{fig:system1_and_system2_timeline}
\end{center}
\vskip -0.2in
\end{figure*}

\subsection{Preliminary: History of System 1 and System 2 in AI}
\label{sec:framework_sys1_and_2}

When confronted with a problem, the human brain can operate in one of two ways: quickly and intuitively, or slowly and deliberately. 
These two processing methods are referred to as System 1 and System 2, or, as Nobel Prize-winning psychologist Daniel Kahneman describes them, ``Fast and Slow Thinking''~\citep{daniel2017thinking}.
System 1 operates quickly and intuitively, making it well-suited for simple and repetitive tasks. In contrast, System 2 functions in a slow and deliberate manner, tailored for complex problem-solving.
There has been a long history of attempts to integrate this dual-process theory into AI systems, as depicted in Figure~\ref{fig:system1_and_system2_timeline}.

Prior to the advent of LLMs, most AI models were characterized by shallow perception and pattern recognition, but lacked capabilities in high-level reasoning, planning, and generalization~\citep{tran2021deep,goyal2022inductive}.
Drawing inspiration from the ``Fast and Slow Thinking'' paradigm, researchers have developed a range of models and frameworks integrating System 1 and 2 abilities tailored to different application scenarios, including reasoning tasks~\citep{hua2022system,nye2021improving}, visual question answering~\citep{liu2022fact, ma2023hybridprompt}, lifelong learning~\citep{pham2021dualnet}, reinforcement learning~\citep{gulati2021interleaving}, among others.
With the advancement of deep learning technologies, the integration of System 1 and System 2 continues to evolve.
During the initial phase of deep learning, machine learning models~\citep{posner2020robots} and shallow neural networks~\citep{anthony2017thinking,ding2019cognitive} were designed primarily for straightforward classification and regression tasks, aligning well with the characteristics of System 1. 
Concurrently, representation learning, exemplified by models like BERT and GPT, also serves as a typical embodiment of System 1 within pre-trained language models~\citep{lugosch2020surprisal,nye2021improving}.
System 2 primarily involves symbolic~\citep{nye2021improving,wu2023symbol} and logical computation~\citep{chen2019deep,ding2019cognitive,hua2022system,ozturkler2022thinksum}, wherein search algorithms such as Monte Carlo methods~\citep{anthony2017thinking,gulati2021interleaving}, Tree Search~\citep{yao2023tree} and Graph Neural Networks~\citep{lugosch2020surprisal,liu2022fact} also function as System 2 models.
Moreover, by incorporating structural bias via the GFlowNet frameworks \citep{bengio2023gflownet, hu2023gflownet}, the system can more effectively support the reasoning, design, and optimization processes of System 2, which, in turn, further strengthens the causal properties of System 2.

In the era of large language models, LLMs exhibiting emergent abilities have the potential to handle complex reasoning tasks beyond mere pattern recognition, embodying the characteristics of System 2~\citep{hagendorff2023human}.
Recent observations and research indicate that LLMs such as ChatGPT and GPT-4, endowed with robust emergent abilities, possess the capability to be classified as System 2 in specific tasks.
\citet{lin2023swiftsage} introduced SwiftSage for multi-step reasoning tasks, wherein a fine-tuned T5 model, obtained through imitation learning, is considered as System 1 for simpler steps, and prompting-based GPT-4 is employed as System 2 for more complex steps, as determined by heuristic rules.
Some studies also classify LLMs as System 2, while smaller models are categorized as System 1~\citep{qi2024interactive,zhang2024fast,zhang2024cogenesis}.
On the other hand, LLMs equipped with specialized reasoning strategies, including Chain-of-Thought~\citep{wei2022chain}, Tree-of-Thought~\citep{yao2023tree}, and Algorithm of Thought~\citep{sel2023algorithm}, and enhanced with self-reflection~\citep{shinn2023reflexion} or debugging capabilities~\citep{chen2023teaching}, demonstrate the ability to solve complex tasks.
Overall, the shortcomings of LLMs in System 2 are obvious, and they have drawn the attention of researchers.
However, the optimal integration of System 1 and System 2 in LLMs remains an area of ongoing exploration. We will explore this topic further in subsequent sections.

\subsection{Layer 1: Foundational Abilities}
\label{sec:framework_layer1}

\subsubsection{Component 1: 
System 1 Capabilities Enhancement
}
\label{sec:framework_layer1_component2}

\hypbox{Takeaway Message}{
System 1, built upon a world knowledge base, should continually be augmented with highly abstracted rules derived from System 2.
Through repetitive experiences of System 2 reasoning, these rules can be transformed into intuitive System 1 abilities, which, in turn, can enhance System 2 capabilities.
}

\begin{figure*}[htbp]
\begin{center}
\centerline{
\includegraphics[width=0.88\textwidth]{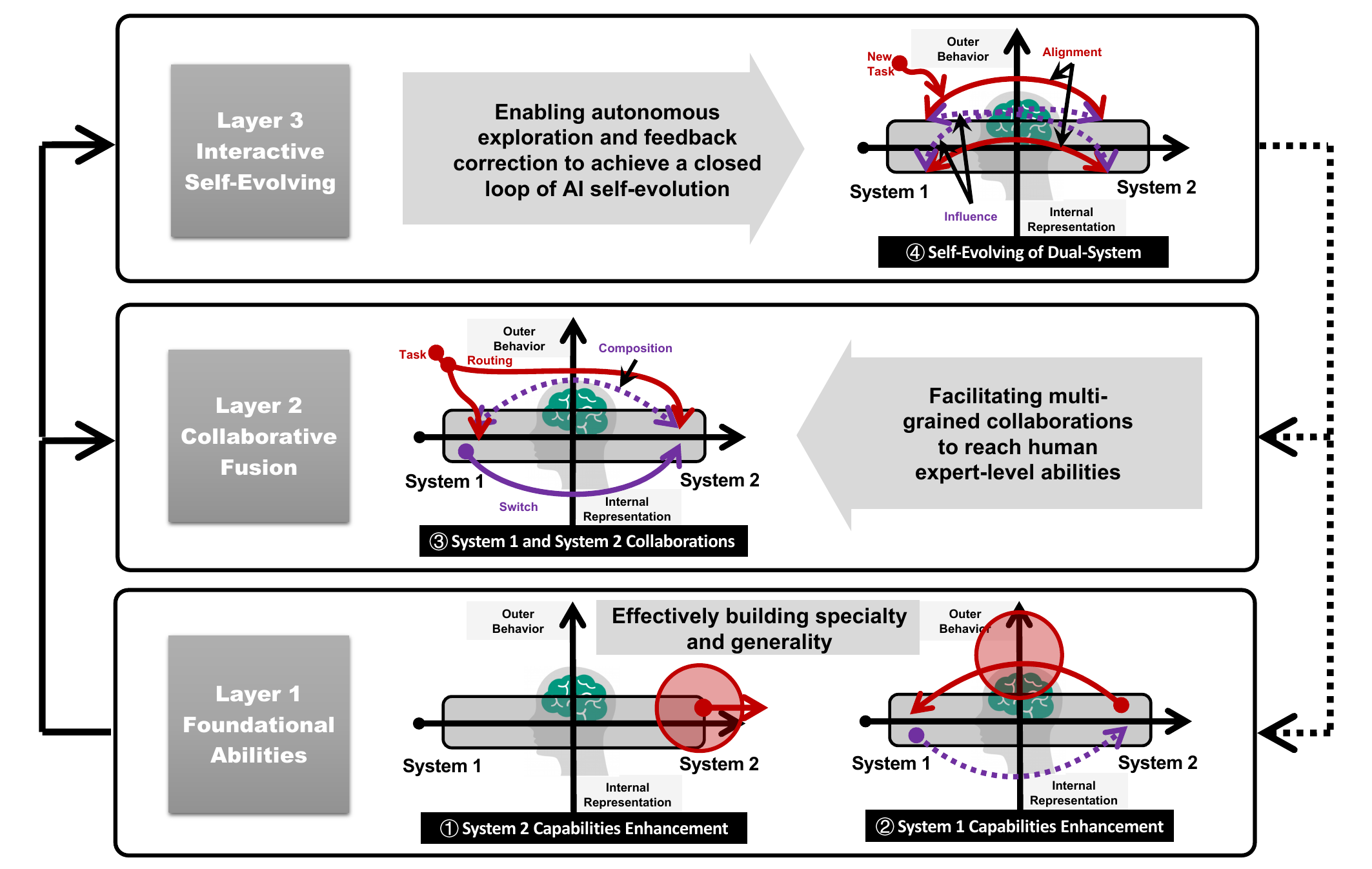}
}
\caption{
The three layers and four key components of our proposed theoretical framework for building specialized generalists from a System 1 and System 2 fusion perspective include the development of both systems (\ding{202} and \ding{203}), their collaboration (\ding{204}), and the self-evolving of dual-system (\ding{205}).
The x-axis represents the two systems, trending more towards System 2 as slow, rational thinking increases.
The y-axis represents potential collaborations involving internal representations and external behaviors among the systems, with human readability improving as collaborations shift from representations to behaviors.
}
\label{fig:Sys1_and_2_four_keys}
\end{center}
\end{figure*}

A well-developed System 2 can stably produce superior answers for given problems, yet it incurs substantial costs during inference and reasoning~\citep{kahneman2003perspective,evans2013dual}. In human cognition, prolonged exercise of System 2 reasoning abilities leads to the abstraction of rules, which then evolve into intuitive responses~\citep{fitts1967human,logan1988toward}. This intuitive ability is evident among experts in specific fields and tasks. 
Therefore, the inherent growth of capabilities in System 1 will naturally arise from the long-term interactive feedback and consolidation provided by System 2.

When we regard LLMs as (weak or approximate) System 2 entities, numerous studies focus on distilling knowledge or rules from larger models into smaller ones~\citep{zhu2023large,cai2023large,zhu2023pad,fu2023specializing}.
This process, known as imitation learning or behavior cloning~\citep{torabi2018behavioral,shao2024beyond}, can be optimized through best-of-N (BoN)~\citep{gao2023scaling} or rejected sampling~\citep{liu2023statistical} techniques.
Subsequently, the System 1 models derived in this way are closely aligned with System 2 models, thereby enhancing their collaborative and problem-solving capabilities. 
However, as the capabilities of System 2 models increase, there remains a need for further exploration to stably extract knowledge.

Ultimately, deriving System 1 models from System 2 models is crucial within the fusion paradigm of Systems 1 and 2.
Fundamentally, System 1 models are highly specialized derivations of System 2 models, designed for specific skills or tasks.
System 1 excels in processing these tasks and exhibit greater stability and efficiency than System 2.
Furthermore, this emergence can also enhance the flexibility of cooperation between systems.

\subsubsection{Component 2: System 2 Capabilities Enhancement}
\label{sec:framework_layer1_component1}

\hypbox{Takeaway Message}{
System 2 capability is essential for acquiring expert-level skills in specific domains and is crucial for achieving compositional generalization.
The scaling of iterative search, combined with self-feedback in the output space, represents a promising approach to enhance System 2 capabilities.
}

Current Large Language Models such as ChatGPT and GPT-4 have demonstrated substantial potential in fulfilling System 2 functions, particularly in emergent capabilities such as reasoning and multi-hop planning~\citep{wei2022chain,yao2023tree}. These models exhibit human-like intuitive behavior and largely eliminate reasoning biases~\citep{hagendorff2023human}.
However, these models are conditionally categorized as System 2, as their capabilities are not yet stable and require further development.

Let’s examine prior Expert ANIs like AlphaGo and AlphaZero~\citep{silver2017mastering}, where the key factor in their success was scaling search.
Scaling search for next token predictions in LLMs represents a highly promising direction for enhancing System 2 abilities, going beyond traditional ``Scaling Laws'' in parameters~\citep{kaplan2020scaling} or training data~\citep{sorscher2022beyond}.
Previously introduced prompt engineering methods~\citep{sahoo2024systematic} are relatively superficial approaches to achieving this goal. Moreover, massive sampling and rewarding of next tokens, terminals, or steps can also be used to augment search capabilities.
In fact, all these search methods and prompts are designed to enhance the working memory of LLMs~\citep{bubeck2023sparks,gong2024working}, fostering more deliberate and slow thinking.
Recently, there has been a trend~\citep{zhang2024rest,tian2024toward,brandfonbrener2024vermcts,zhang2024accessing} to augment LLMs with search algorithms like Monte Carlo Tree Search~\citep{swiechowski2023monte} and Q*~\citep{mcintosh2023google}.
However, these efforts primarily focus on math or coding tasks that possess ground truths.
The implementation of scaling search in open-ended generation~\citep{chi2024thoughtsculpt} remains largely unexplored.
Additionally, researchers are investigating the integration of formal logic, such as neuro-symbolic systems~\citep{trinh2024solving}, to facilitate more meticulous, logic-based decisions. However, this approach still faces significant challenges in terms of flexibility and universality. Moreover, budget-aware scaling must also be considered~\citep{wang2024reasoning}.
Furthermore, GFlowNet \citep{bengio2023gflownet, hu2023gflownet} can further reinforce structural priors through implicit search behaviors, effectively boosting diversity. However, these frameworks are still subject to limitations in efficiency and stability.

Overall, enhancing System 2 abilities is crucial for developing specialized generalists capable of achieving expert-level performance in targeted domains. Optimal methods for this enhancement still require further investigation. Among these, scaling search emerges as a practical and promising approach worth exploring.

\subsection{Layer 2: Collaborative Fusion}
\label{sec:framework_layer2}

\subsubsection{Component 3: System 1 and 2 Collaborations}
\label{sec:framework_component3}

\hypbox{Takeaway Message}{
Collaborations between Systems 1 and 2 can strike a balance between effective information processing and high-quality decision-making. These collaborations also exist within each individual system. This balance further enhances adaptation to new environments and improves resource allocations.}

Once System 1 and System 2 models are established, the next crucial step involves leveraging their respective strengths to address incoming tasks. Mirroring human cognition, System 1 initially excels at solving given problems, while System 2 corrects biases as needed~\citep{kahneman1977intuituve}. These basic collaborations, although fundamental, present challenges in dynamically scheduling System 1 and System 2 interactions in real-world environments~\citep{stanovich2000advancing,evans2003two}.

Regarding LLMs, the internal working mechanisms within these models remain largely unknown, particularly the specific functionalities activated in transformer-like circuits~\citep{elhage2021mathematical,olsson2022context}.
Therefore, managing or controlling collaborations within these black-box LLMs is challenging.
Collaborations at the representation level between models are rarely explored; however, research on relative representations~\citep{moschella2023relativerepresentationsenablezeroshot} and representation alignment~\citep{sucholutsky2023getting} may offer potential methods for enhancing internal collaborations.
On the other hand, related research primarily focuses on external collaborations mainly with language behaviors, like model routing~\citep{shnitzer2023large,ding2024hybrid}, cascade~\citep{dohan2022language,yue2023large} or selection~\citep{zhao2023automatic} based on priors regarding candidates or model uncertainty~\citep{xiong2023can}.
Further studies also investigate advanced task decomposition or real-time planning~\citep{yao2022react,yao2023tree,shinn2023reflexion}, primarily within LLM-based agents~\citep{wang2023survey}.
More detailed investigations into collaborations involve models’ decoding processes like token predictions or step generations~\citep{zhang2024fast} besides of above sample-level.
These are explored in techniques such as speculative decoding~\citep{leviathan2023fast}, contrastive decoding~\citep{li2022contrastive,o2023contrastive}, or proxy tuning~\citep{mitchell2023emulator,liu2024tuning} within collaborative frameworks between large and small language models.
Overall, numerous studies have addressed planning and decoding in LLMs. However, future research inspired by human cognitive processes or alternative scheduling methods for System 1 and System 2 remains a promising area of exploration. Particularly, the step-level collaborations between the two systems during the generation of free-form responses warrant further investigation.

Broadly speaking, the principle ``More is different'' applies well in the human brain and complex systems design~\citep{anderson1972more}, which merge collaborative intelligence. Systems 1 and 2 should not be confined to a single model but should be expanded to include multiple and even massive models. Therefore, collaborations occur not only between Systems 1 and 2 but also within the internal models of each system. These represent complex collaborative behaviors that are still under exploration in the era of LLMs.

\subsection{Layer 3: Interactive Self-Evolving}
\label{sec:framework_layer3}

\subsubsection{Component 4: Self-Evolving of Dual-System}
\label{sec:framework_component4}

\hypbox{Takeaway Message}{
Systems 1 and 2 can evolve through dual-system interactions and self-evaluations on dynamic, unseen tasks. Moreover, the dual-system can also achieve evolution through interactions with the physical world by conducting automatic discoveries using virtual simulations and external tools.
}

The interaction between Systems 1 and 2 is dynamic and continually evolves in response to a changing environment and emerging tasks and challenges.
This ongoing adaptation necessitates continuous learning and development of both systems~\citep{stanovich2000advancing,buss2019evolutionary}.

Unlike previous studies that integrate the concepts of Systems 1 and 2 within continual learning frameworks \citep{pham2021dualnet,arani2022learning,wang2023comprehensive}, this component focuses on addressing the unique continual learning challenges faced by each model.
It aims to perpetually enhance the capabilities of both System 1 and System 2 models, thus improving their combined efficiency and advancing their applications.
While Systems 1 and 2 are considered within the context of LLMs, established techniques to combat ``catastrophic forgetting'' \citep{kirkpatrick2017overcoming,de2021continual} are still relevant for achieving lifelong learning goals.
However, a new challenge within the dual-process framework is ensuring consistent alignment for prolonged collaboration between the two systems, including both behavior~\citep{gupta2024behavior} and representation~\citep{sucholutsky2023getting}.
This alignment is analogous to aligning LLMs with human intentions, values, and the principles of being helpful, honest, and harmless (3H)~\citep{bai2022training,wang2023aligning}, aiming for seamless integration between the systems.
Recent studies investigate updating knowledge in models through modular approaches that mimic the dynamic memory mechanisms in the brain~\citep{wang2023incorporating}. In terms of Systems 1 and 2, it is crucial to maintain stability in one system while updating the other, ensuring consistency throughout the process~\citep{qi2024interactive}. One strategy involves modular knowledge storage and management~\citep{qi2024online}.
In addition to interactive continual learning from environment, it is also crucial for systems to self-evolve~\citep{jiang2023selfevolve,yuan2024self,tao2024survey} using synthetic data generated by the LLMs themselves within virtual tools~\citep{bauer2024comprehensive,liu2024best}.

In conclusion, defining the alignment between the dual processes is challenging but crucial for continual task learning.
For specialized generalists, maintaining the ability to learn continually and process new tasks is essential.
During this period, modular knowledge transfer between Systems 1 and 2 is vital to ensure the entire system operates in a closed loop~\citep{stanovich2000advancing,floreano2008bio}.

\section{Challenges and Future Directions}
\label{sec:challenge}

\subsection{Building Models Collaboration Laws.}
\label{sec:challenge_scaling_law}
The performance improvement of current LLMs heavily relies on ``Scaling Laws''~\citep{kaplan2020scaling,hoffmann2022training}, where an increase in parameters and training data generally leads to predictable more powerful performance.
However, there is considerable debate regarding the upper limits of Scaling Laws due to the potential for exhausting available web data~\citep{villalobosposition} and limited computation resource.
How can we build upon scaling laws by models or systems interactions? It is crucial to implement the fusion of massive specialized and general models, which may provide new avenues to further enhance the effectiveness of scaling laws at the model level.
The ``Scaling laws'' arising from the coordination among intelligent agents are being explored~\citep{qian2024scaling,wang2024mixture,li2024more}, demonstrating promising directions for further scaling based on traditional scaling laws.
Based on interactive media, these collaborations can occur during various periods, such as at the parameter level through model merging~\citep{wortsman2022model,stoica2023zipit,goddard2024arcee,akiba2024evolutionary}, at the representation level through representation engineering~\citep{sucholutsky2023getting,zou2023representation}, at the logit level through inference-time alignment~\citep{huang2024deal,wan2024knowledge,ding2024mastering,zhang2024fast}, and at the token~\citep{jin2024collaborative} or language level~\citep{guo2024large}, which are the most common scenarios in various multi-agent collaborations.
Recent work show the collaborations between large and small models seem to be predicatble, further explorations on multiple models collaborations can be studied~\citep{zhang2024fast}, which is a potential way to augment current LLMs and to achieve more powerful specialized generalists beyond ``Scaling Law''.

\subsection{Building Data Mixture Laws.}
\label{sec:challenge_data_law}

Beyond the development of specialized generalists through model-level collaborations, data plays a crucial role in achieving specialized generalist capabilities. Various data-centric methods~\citep{zha2023data} can be effectively utilized to reach this goal.
For instance, high-quality and diverse domain-specific pre-training data~\citep{penedo2024fineweb,liu2024best,longpre2023pretrainer} or instructions~\citep{ding2023enhancing,liu2023makes} are essential for acquiring domain knowledge.
Simultaneously, studying the mixture of domain data is important, as it helps balance speciality and generality~\citep{Zhang2024UltraMedicalBS}.
While there appear to be underlying principles in data mixture~\citep{xie2024doremi,ye2024data}, these still require further practical analysis and experimentation. In summary, inspired by data pruning laws~\citep{sorscher2022beyond}, data mixture could also be established as a predictive law to guide our efforts in developing specialized generalist AI.

\subsection{New Evaluation and Benchmarks.}
\label{sec:challenge_eval}
As LLM capabilities improve, traditional benchmarks~\citep{hendrycks2020measuring,cobbe2021training,hendrycks2021measuring,chen2021evaluating} are nearing saturation, making it increasingly difficult to discern differences between models.
In response, researchers are developing more challenging benchmarks that use human experts as the standard, such as GPQA~\citep{rein2023gpqa} and AGIEval~\citep{zhong2023agieval}, or employ real-world use cases for testing, such as SWE~\citep{jimenez2024swebench} and BigCodeBench~\citep{zhuo2024bigcodebench}.
These benchmarks encompass a variety of specific domain problems. However, the diversity and quantity of these problems are still insufficient to fully validate the specialization abilities within domains.
Additionally, benchmarks designed to assess the impact of generalization on specialization are under exploration. Such benchmarks are crucial for validating the effectiveness of models and data mixture methods in balancing specialty and generality.

\subsection{New Architecture from Scratch.}
\label{sec:challenge_archi}
Recent work has demonstrated that designs inspired by human-like networks can provide superior systematic generalization~\citep{lake2023human}. Due to the opaque nature of LLMs, replicating human cognitive processes remains unfeasible. Beyond traditional transformers (including various implementations and linear variants like Mamba~\citep{gu2023mamba}) and autoregressive learning, exploring new architectures is crucial. Such architectures include GFlowNet~\citep{lahlou2023theory,jain2023gflownets,bengio2023gflownet} and KAN~\citep{liu2024kan,bozorgasl2024wav}, which hold significant potential for enhanced reasoning and controllability.
Recent work has also begun to combine these models to create hybrid architectures~\citep{lieber2024jamba,qi2024smr}.
Additionally, designing new architectures inspired by human cognitive systems like Systems 1 and 2 is a promising avenue.
However, further efforts are necessary to determine whether these architectures can scale or surpass LLMs in innovative ways.
In addition, some research has explored new architectural designs from the perspective of decoupling memory and reasoning \citep{chen2022decoupling, qi2024interactive, qiao2024prism}. These works have focused on designing externally pluggable memory components \citep{qi2024interactive, qi2024contrastive} and modular architectural learning strategies. However, there is limited research on constructing more compositional model designs that can simultaneously enable efficient memory storage and sustainable learning capabilities.

\subsection{Applications in Multi-modal and Embodied AI}
As indicated in the discussion on enhancing System 2 abilities, increasing the intelligence density of supervised signals is a promising trend.
This approach has proven useful in multi-modal pre-training~\citep{Chen_2024_CVPR,ma2022cmal,bai2023qwen}, as multi-modal data can provide more fine-grained and diverse information than text alone~\citep{ma2022unitranser}. Moreover, token alignment and pre-training between across-modality information may be the key to improve the intelligence density of supervision signals and further realizing the emergence of foundational models and AGI.
In this regard, it is necessary to extend research and applications of specialized generalists to multi-modal and embodied AI. Embodied intelligence~\citep{gupta2021embodied} is an another effective means to achieve general and specialized integration, and is also the most likely path to building AGI that understands the physical world. However, embodied intelligence is not just the application of Multi-modal LLMs plus robots, whereas requires the timely self-evolution according to the feedback from the physical world. In a multi-modal physical world, AI systems should possess the potential to learn from the environment and receive feedback that facilitates self-evolution.

\subsection{Applications in Scientific Discovery}
\label{sec:challenge_scientific}

Scientific discovery in biomedicine~\citep{qi2023large}, chemistry~\citep{m2024augmenting}, physics~\citep{ma2024llm}, mathematics~\citep{romera2024mathematical}, and related domains drives advancements in human societal development.
Recent work~\citep{park2023papers} indicates that research innovation has slowed due to the emergence of massive literature within specific domains, resulting in an ``information cocoon.''~\citep{sunstein2006infotopia}
Traditional AI-based scientific tools designed for specific tasks, such as linear classifiers and graph neural networks, lack the generality needed to break out of these information cocoons.
LLM-based scientific tools~\citep{ai4science2023impact,zhang2024scientific,liang2024mapping} introduce new possibilities for navigating vast networks of literature by internalizing all knowledge from the corpus available on the internet.
These tools open up a new pathway that simulates real-world researchers~\citep{m2024augmenting}, encompassing the uncertainty inherent in exploring unseen fields.
However, due to the limited generality of LLMs in specific domains, specializing in particular areas has become a popular method to enhance specificity~\citep{yue2023mammoth,zhu2023pad,bolton2024biomedlm}.
Balancing specificity and generality in scientific research fields is expected to unlock further potential~\citep{zhang2024chemllm,Zhang2024UltraMedicalBS}.
The challenge of building specialized generalists for scientific discovery continues to be a critical area of exploration.

\subsection{Risks and Controllability.}
\label{sec:challenge_risk}
Although Specialized Generalist Intelligence is not as advanced as AGI, there are still risks associated with generating harmful content~\citep{weidinger2022taxonomy,weidinger2021ethical,anil2024many}. The factuality of generated content must also be monitored, with particular attention to the study of hallucination phenomena~\citep{xu2024hallucination,huang2023survey,jiang2024large}.
For expert-level applications such as clinical surgery~\citep{Zhang2024UltraMedicalBS} and financial analysis~\citep{kim2024financial}, it is crucial that the outputs of models are highly controllable~\citep{10.1145/3555803,sun2024trustllm}.
Relatively speaking, the boundaries of specialized generalists are clearer than those of generalists, which makes them more controllable.
This controllability depends on the specialization of System 1 and the fault tolerance of System 2.
Compared to solely pursuing performance improvement or security enhancement, we need to adopt a more balanced strategy that comprehensively considers and coordinates their simultaneous development. This approach ensures that performance and security enhancements do not compromise each other. Moreover, it adheres to an AI-$45^\circ$ law, which further drives the sustainable and healthy development of the entire system.

\section{Conclusion}
\label{sec:conclusion}
In this paper, we formally propose the concept of Specialized Generalist AI (SGI), a pivotal milestone in transitioning from current LLMs to Expert AGI. The purpose of SGI is to effectively enhance specialized capabilities while simultaneously developing general capabilities in a balanced manner. This strategic approach allows for rapid advancement into high-value areas and fosters the creation of an efficient data feedback loop, thereby accelerating the development of AGI.
SGI is delineated into three stages, determined by the number of specialized skills and the level of general performance. We introduce a conceptual framework for implementing SGI, viewed through the lens of Systems 1 and 2, and detail the four key components of this framework. Finally, we outline the challenges and potential directions for furthering SGI development.

\newpage

\addcontentsline{toc}{section}{Contributions}
\section*{Contributions}
\label{sec:contributions}
The contributions of this paper are outlined as follows:
\begin{itemize}
    \item Bowen initially proposed the concept of Specialized Generalist AI (SGI) and the technical roadmap from the perspective of Systems 1 and 2.
    \item Kaiyan was primarily responsible for leading the discussion, refining the levels of SGI, and writing the initial draft. Together with Biqing, Kaiyan refined the initial concept and framework, which includes the three layers and four components necessary for achieving SGI.
    \item All authors participated actively in discussions and the polishing of the paper.
\end{itemize}

\addcontentsline{toc}{section}{Acknowledgements}
\section*{Acknowledgements}
\label{sec:acknowledgements}
Professor Bowen Zhou has delivered numerous presentations on Specialized Generalists and Systems 1 and 2 over the past few years at various forums and talks. We thank everyone for their valuable feedback and discussions on these topics with experts and scholars from different domains, including Daniel Kahneman, Andrew Chi-Chih Yao, Thomas J Sargent, and others.
We also express our gratitude to Ning Ding, Zhiyuan Ma, Xuekai Zhu, and Ermo Hua in Center for Collaborative \& Conversational Intelligence at Tsinghua University for their helpful advice on writing and content.
This work is supported by the National Science and Technology Major Project (2023ZD0121403).

\newpage
\bibliographystyle{text}
\bibliography{main,levels_of_AGI,sys1_and_2}

\newpage
\appendix

\clearpage

\end{document}